\pdfoutput=1
\RequirePackage{amsmath}

\documentclass[runningheads]{llncs}

\usepackage[dvipdfmx]{graphicx}

\usepackage{adjustbox}
\usepackage{amssymb}
\usepackage{animate}
\usepackage{bmpsize}
\usepackage{color}
\usepackage{comment}
\usepackage{cite}
\usepackage{float}
\usepackage[hidelinks]{hyperref}
\usepackage{multirow}
\usepackage[caption=false]{subfig}
\usepackage{textcomp}
\usepackage{todonotes}
\usepackage{xcolor}

\begin{document}

\title{INSIDE: Steering Spatial Attention with Non-Imaging Information in CNNs}
\titlerunning{INSIDE}

\author{Grzegorz Jacenk\'ow\inst{1} \and
Alison Q. O'Neil\inst{1,2} \and
Brian Mohr\inst{2} \and
Sotirios A. Tsaftaris\inst{1,2,3}
}

\authorrunning{G. Jacenk\'ow et al.}

\institute{The University of Edinburgh, United Kingdom \\
          \email{g.jacenkow@ed.ac.uk} \and
          Canon Medical Research Europe, United Kingdom \and
          The Alan Turing Institute, United Kingdom
}

\maketitle

\begin{abstract}
We consider the problem of integrating non-imaging information into segmentation
networks to improve performance. Conditioning layers such as FiLM provide the
means to selectively amplify or suppress the contribution of different feature
maps in a linear fashion. However, spatial dependency is difficult to learn
within a convolutional paradigm. In this paper, we propose a mechanism to allow
for spatial localisation conditioned on non-imaging information, using a
feature-wise attention mechanism comprising a differentiable parametrised
function (e.g. Gaussian), prior to applying the feature-wise modulation. We name
our method INstance modulation with SpatIal DEpendency (INSIDE). The
conditioning information might comprise any factors that relate to spatial or
spatio-temporal information such as lesion location, size, and cardiac cycle
phase. Our method can be trained end-to-end and does not require additional
supervision. We evaluate the method on two datasets: a new CLEVR-Seg dataset
where we segment objects based on location, and the ACDC dataset conditioned on
cardiac phase and slice location within the volume. Code and the CLEVR-Seg
dataset are available at \url{https://github.com/jacenkow/inside}.

\end{abstract}

\section{Introduction}
Acquisition of medical images often involves capturing non-imaging information
such as image and patient metadata which are a source of valuable information
yet are frequently disregarded in automatic segmentation and classification. The
useful information should expose correlation with the task such as body mass
index (BMI) with ventricular volume~\cite{bai2018automated}, or symptom
laterality with stroke lesion laterality~\cite{sato2012conjugate}, and these
correlations can be exploited to improve the quality of the structure
segmentation. Nevertheless, combining both imaging and non-imaging information in
the medical domain remains challenging, with dedicated workshops to approach
this problem~\cite{stoyanov2018graphs}.

\begin{figure}
  \centering
  \includegraphics[width=\textwidth]{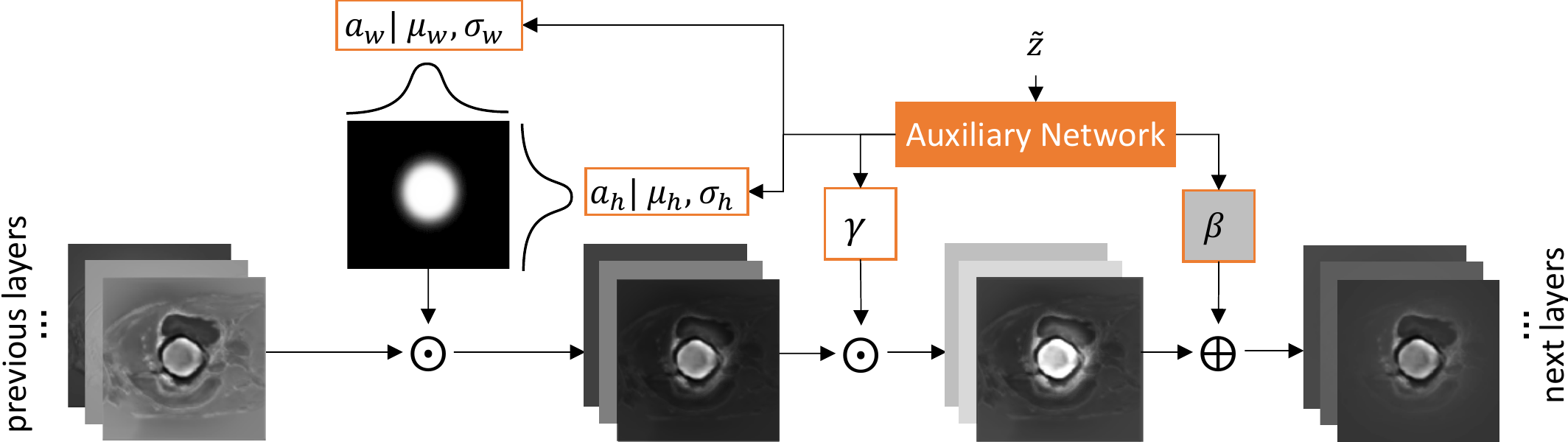}
  \caption{Visualisation of the method. Given a feature map $F_c$ and conditioning
           vector $\tilde{z}$, the method first applies spatial attention ($a$) with
           scale ($\gamma$) and shift ($\beta$) factors to $F_c$ respectively.
           The attention matrix ($a$) is the product of two Gaussian vectors
           ($a_h, a_w$). Therefore, for a single feature map, the auxiliary network
           predicts six parameters ($\gamma$, $\beta$, $\mu_h$, $\sigma_h$, $\mu_w$,
           $\sigma_w$). We denote Hadamard product with $\odot$ symbol.}
  \label{fig:method}
\end{figure}

Conditioning layers have become the dominant method to tackle this challenge,
finding application in image synthesis~\cite{brock2018large}, style
transfer~\cite{huang2017arbitrary} and visual question answering
(VQA)~\cite{perez2018film}. In this setup, the network is conditioned on
non-imaging information via a learned set of scalar weights which affinely
transform feature maps to selectively amplify or suppress each feature, thus
controlling its contribution to the final prediction. However, this method has
limited capability to adjust channels spatially, and is less suited to
conditioning on information relating to spatial or spatio-temporal prior
knowledge. Consider a problem where we expect to produce a segmentation only on
one side of the image (left or right) indicated by the laterality of the
patient's symptoms. To accomplish this task, the network would require to learn
how to encode relative spatial relationships and split them into channels. We
show that spatial conditioning can be challenging and propose a method to
overcome this limitation.

We present a new conditioning layer which uses non-imaging information to steer
spatial attention before applying the affine transformation. We choose a
Gaussian for the attention mechanism due to its parameter-efficiency, allowing
us to learn a separate attention per channel. However, other differentiable
functions can also be used. We first test our method on a simulated dataset, our
extension of the CLEVR\footnote{Diagnostic Dataset for Compositional Language
and Elementary Visual Reasoning} dataset~\cite{johnson2017clevr}, where we
segment objects based on their location within the image space. To prove the
method is applicable in a clinical setting, we use the ACDC\footnote{Automated
Cardiac Diagnosis Challenge (ACDC), MICCAI Challenge
2017} dataset~\cite{bernard2018deep} with the task to segment anatomical
structures from cardiac cine-MR images. We perform 2D segmentation, and provide
slice position and cardiac cycle phase as the non-imaging information to our
method.

\textbf{Contributions}: \textbf{(1)} we propose a new
conditioning layer capable of handling spatial and spatio-temporal dependency
given a conditioning variable; \textbf{(2)} we extend the CLEVR dataset for
segmentation tasks and several conditioning scenarios, such as shape-,
colour-, or size-based conditioning in the segmentation space;
\textbf{(3)} we evaluate different conditioning layers for the task of
segmentation on the CLEVR-Seg and ACDC datasets.

\section{Related Work}
\label{sec:related}
An early work on adapting batch normalisation for conditioning was in style
transfer. The conditional instance normalisation
layer~\cite{dumoulin2016learned} (Eq.~\ref{eq:instance}) applied a pair of scale
($\gamma_s$) and shift ($\beta_s$) vectors from the style-dependent parameter
matrices, where each pair corresponded to a single style $s$ such as Claude
Monet or Edvard Munch. This allowed several styles to be learned using a single
network and proved that affine transformations were sufficient for the task.
However, the method is restricted to the discrete set of styles seen during
training. In Adaptive Instance Normalisation (AdaIN)~\cite{huang2017arbitrary},
the authors proposed to instead use a network to predict the style-dependent
vectors (as in hypernetworks), allowing parameters to be predicted for arbitrary
new styles at inference time.
\begin{equation}
  z = \gamma_s \frac{x - \mu_x}{\sigma_x} + \beta_s
\label{eq:instance}
\end{equation}

AdaIN has been applied outside of the style transfer domain, for instance to
image synthesis using face landmarks where the method is used to inpaint the
landmark with face texture~\cite{zakharov2019few}, and to conditional object
segmentation given its coordinates~\cite{sofiiuk2019adaptis}. A similar method
to AdaIN was applied to visual question-answering (VQA); the authors used
feature-wise linear modulation layer (FiLM)~\cite{perez2018film} to condition
the network with questions. FiLM is identical to AdaIN but omits the instance
\emph{normalisation} step ($\mu_x, \sigma_x$ in Eq.~\ref{eq:instance}), which
the authors found to be unnecessary. FiLM has found application in medical image
analysis for disentangled representation
learning~\cite{chartsias2019disentangled} and for
segmentation~\cite{jacenkow2019conditioning}.

A drawback of both AdaIN and FiLM is that they manipulate whole feature maps in
an affine fashion, making the methods insensitive to spatial processing. To
overcome this limitation, SPADE~\cite{park2019semantic} was proposed, where a
segmentation mask is used as a conditioning input in the task of image
synthesis, leading to both feature-wise and class-wise scale and shift
parameters at each layer. This method is not suitable if the non-imaging
information cannot be conveniently expressed in image space.

The closest method to ours is \cite{rupprecht2018guide}. The authors proposed to
extend FiLM with spatial attention, creating a Guiding Block layer, in which the
spatial attention is defined as two vectors $\alpha \in \mathbb{R}^H$ and $\beta
\in \mathbb{R}^W$ which are replicated over the $H$ and $W$ axes and added to
the global scale factor ($\gamma_c^{(s)}$) as shown in Eq.~\ref{eg:guideme} (the
authors call the shifting factor as $\gamma_c^{(b)}$). This spatial conditioning
is expensive as there are an additional $H + W$ parameters to learn; perhaps for
this reason, a single attention mechanism is learned for each layer and applied
across all feature maps.

\begin{equation}
  F'_{h, w, c} = (1 + \alpha_h + \beta_w + \gamma_c^{(s)}) F_{h, w, c} + \gamma_c^{(b)}
\label{eg:guideme}
\end{equation}

\noindent
In our work, we utilise a learned attention mechanism for each feature map. Our
mechanism is similar to \cite{kosiorek2017hierarchical}, where the product of
two Gaussian matrices parametrised by mean ($\mu$), standard deviation
($\sigma$) and stride ($\gamma$) between consecutive Gaussians (one Gaussian per
row, one matrix per axis) is constructed. However, the relation between standard
deviations and strides is estimated before the training and kept fixed. Our
method applies a single Gaussian vector per axis (no stride) and we train the
whole method end-to-end. Further, the parameters
in~\cite{kosiorek2017hierarchical} are estimated using consecutive input images
whilst we use an auxiliary conditioning input, and we combine with a FiLM layer.

\section{Method}
\subsection{INstance modulation with SpatIal DEpendency (INSIDE)}
Our method adopts the formulation of previous conditioning
layers~\cite{ulyanov2016instance} where, given a feature map $F_c$, we apply an
affine transformation using scale ($\gamma$) and shift ($\beta$) factors.
However, to facilitate spatial manipulation we propose to apply a Gaussian
attention mechanism prior to feature-wise linear modulation (FiLM) to process
only a (spatially) relevant subset of each feature map. The choice of the
attention mechanism, where each matrix is constructed with two Gaussian vectors
[($a_h | \mu_h, \sigma_h$), ($a_w | \mu_w, \sigma_w$)] is motivated by parameter
efficiency. Therefore, the method can learn one attention mechanism per feature
map by adding four additional parameters for each channel (six parameters in
total, including the scale and shift factors).

We illustrate the method in Fig.~\ref{fig:method}. Given a feature map $F_c \in
\mathbb{R}^{H \times W}$ as input, where $c$ is the channel, we first apply
Gaussian attention similar to~\cite{kosiorek2017hierarchical}. We define two
vectors $a_{c, h} \in \mathbb{R}^H$ and $a_{c, w} \in \mathbb{R}^W$ following
the Gaussian distribution parametrised by mean and standard deviation to
construct an attention matrix $a_c = a_{c, h} a_{c, w}^T$. The attention is
applied to the feature map prior to feature-wise modulation, i.e.
\begin{equation}
  \text{INSIDE}(F_c | \gamma_c, \beta_c, a_c) = F_c \odot a_c \odot \gamma_c + \beta_c \text{ .}
  \label{eq:inside}
\end{equation}

To construct the Gaussian vectors, we normalise the coordinate system,
transforming each axis to span the interval $[-1, 1]$. We apply a similar
transformation to the standard deviation; the value (the output of a sigmoid
activation) lies within the $[0, 1]$ range. We set the
maximum width to 3.5 standard deviations to cover (at maximum) 99.95\% of the
image width, thus constraining by design the allowable size of the Gaussian.

\subsection{Auxiliary Network}
We use a separate auxiliary network (a hypernetwork) for each layer to predict
the parameters of INSIDE (see Eq.~\ref{eq:inside}). The network takes a
conditional input $\tilde{z}$ to control the Gaussian attention and affine
transformation. The information is encoded using a 3-layer MLP arranged as
($\frac{c}{2} - \frac{c}{2} - 6c$) where $c$ is the number of channels feeding
into the INSIDE layer. We use $tanh$ activation functions except for the last
layer where scale ($\gamma$) and shift ($\beta$) factors  are predicted with no
activation (identity function). The Gaussian's mean is bounded between $[-1, 1]$
(relative position along the axis from the centre), enforced with a $tanh$
function, and we use $sigmoid$ activations to predict the standard deviation of
the Gaussian vectors.

\subsection{Loss Function}
To avoid the network defaulting to a general solution with a large diffuse
Gaussian~\cite{nibali2018numerical}, we add an $L_2$ regularisation penalty
$\eta$ to the cost function to encourage learning of localisation, as in the
equation below:

\[ \mathcal{L} = \mathcal{L}_{\text{Dice}} + 0.1 \cdot \mathcal{L}_{\text{Focal}} + \eta || \sigma ||_2^2 \text{ .} \]

The first part of the cost function relates to the segmentation task and
involves a combination of Dice loss~\cite{dice1945measures} (evaluated on the
task foreground classes) and Focal loss~\cite{lin2017focal} (evaluated on every
class including background). The second part is the penalty applied to our
conditioning layer. Throughout the training, we keep $\eta = 0.0001$ and the
Focal loss focusing parameter $\gamma = 0.5$. The coefficients were selected
using a grid search giving reasonable performance across all tested
scenarios. We optimise every model with Adam~\cite{kingma2014adam} with learning
rate set to 0.0001, and $\beta_1 = 0.9$, $\beta_2 = 0.999$. We apply early
stopping criterion evaluated on validation set using Dice score only.

\section{Experiments}
\label{sec:experiments}
We evaluate our method on two datasets and report the Dice coefficient on 3-fold
cross validation. Each experiment was further repeated three times using
different seeds to avoid variance due to the weight initialisation. We compare
our method against the following techniques, as discussed earlier in
Section~\ref{sec:related}:

\begin{itemize}
  \item Baseline - a vanilla CNN network without conditioning
  \item FiLM~\cite{perez2018film} - feature-wise affine transformation
        (also component of INSIDE)
  \item Guiding Block~\cite{rupprecht2018guide} - extension of FiLM with spatial attention
\end{itemize}

\subsection{CLEVR-Seg Dataset}
We present a novel dataset based on the CLEVR dataset~\cite{johnson2017clevr},
which we name CLEVR-Seg. We have extended the original dataset
with segmentation masks and per-object attributes, such as
colour (yellow, red, green), location (quadrant containing the centre of mass),
shape (cubes, spheres, prisms), and size (small, medium, large). The attributes
determine the segmentation task, i.e. \textit{segment red objects},
\textit{segment objects in the bottom left quadrant}, etc. The network must thus
use the non-imaging information to produce an accurate result. In contrast to
the original work, conditioning is provided as categorical one-hot encoded
vectors, rather than as natural language questions since VQA is not our primary
focus. We generated 4000 random images with 3 to 5 objects each (containing at
least one of each shape, size and colour), paired with segmentation masks for
which each conditioning factor was drawn at random with equal probability. We
split the dataset into training (2880 samples), validation (320), and test (800)
subsets which we kept fixed throughout the evaluation. The intensities in each
image were normalised to fit the [0, 1] range.

\begin{figure}
  \centering
  \includegraphics[width=\textwidth]{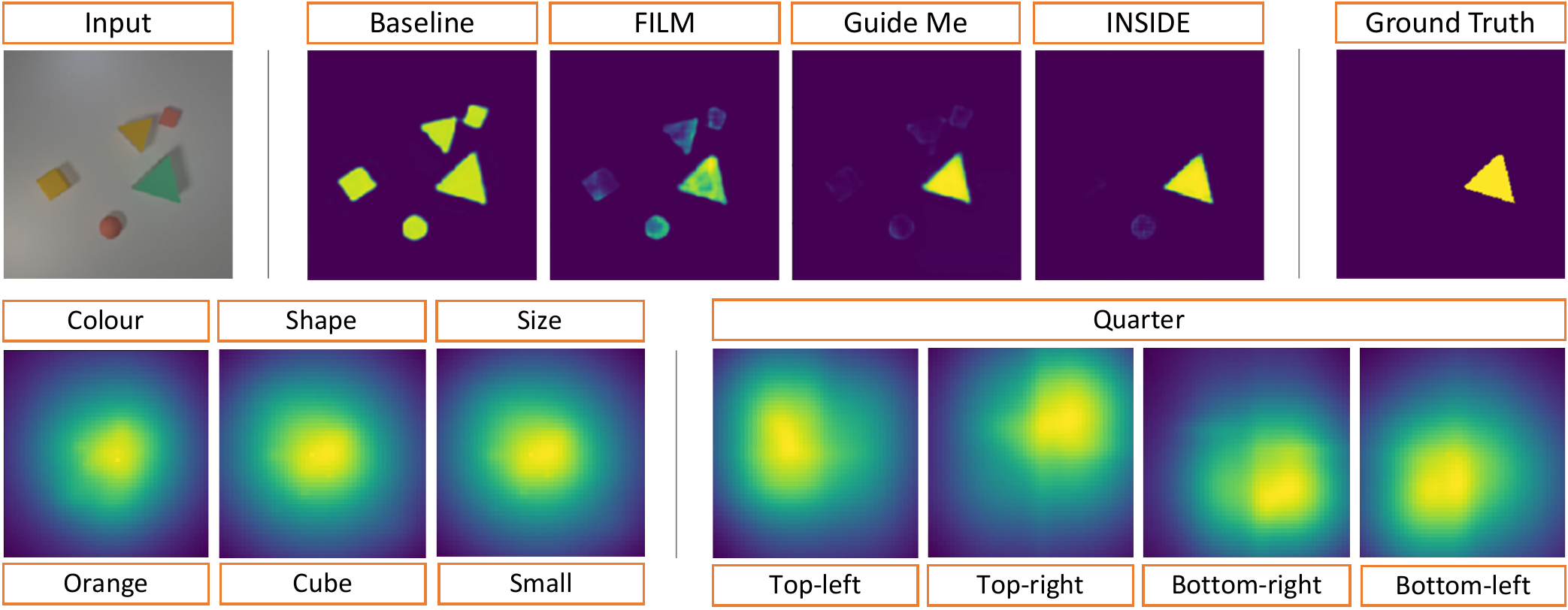}
  \caption{\textbf{Top}: Segmentations produced by the conditioning methods on
           CLEVR-Seg dataset where we condition objects based on the location
           (bottom-right quadrant). \textbf{Bottom}: Learned Gaussians from
           INSIDE (averaged across all feature maps) for different conditioning
           scenarios. The first three scenarios are spatially independent and
           thus, the attentions default to general solutions with diffuse
           Gaussians.}
  \label{fig:clevr-segmentations}
\end{figure}

\subsubsection{Network:} We use a simple fully-convolutional encoder-decoder
with 3 down- and 3 up-sample blocks. Each block consists of ($3 \times 3$)
kernels followed by ReLU activation function and max-pooling/up-sampling,
starting with 16 kernels and doubling/halving at each subsequent step. We test
each conditional layer by placing it between the encoder and the decoder, i.e.
at the network bottleneck.

\subsubsection{Results:} We first evaluate the spatial conditioning scenario.
Quantitative results are shown in Table 1 and qualitative examples are presented
in Fig.~\ref{fig:clevr-segmentations} top. We observe that FiLM has poor
performance, achieving a Dice score of 0.487 ($\pm 0.007$). This result confirms
our hypothesis that spatial conditioning is difficult to disentangle into
separate channels, otherwise FiLM would achieve satisfactory performance. On the
other hand, the Guiding Block achieves adequate performance with a Dice score of
0.819 ($\pm 0.033$). When we use only one Gaussian attention, INSIDE performs
worse than the Guiding Block, however when we use one learned attention per
channel, INSIDE performs best, with a Dice score of 0.857 ($\pm$ 0.025). We
further evaluate our attention mechanisms without feature-wise modulation
(``Single Attention'', ``Multiple Attentions''). These methods yield
satisfactory results, although with higher variance.  The combination of both
gives the highest Dice score and lowest variance across all evaluation
scenarios. The use of feature-wise attention mechanisms give more flexibility to
learn shape- and size-dependent positional bias. Similar patterns are seen on
other conditioning scenarios, i.e. colour, shape and size.

\begin{table}[t]
\label{tab:clevr}
  \begin{center}
    \caption{Quantitative results on our CLEVR-Seg dataset
             where we condition on a quadrant (spatial conditioning), object
             colour, shape and size. We report Dice scores (multiplied by 100)
             with standard deviations shown as subscripts. The method with highest average result
             is presented in \textbf{bold}. We use the Wilcoxon test to assess statistical
             significance between INSIDE (with multiple attentions) and the best baseline method.
             We denote two (**) asterisks for $p \leqslant 0.01$.}
    \begin{adjustbox}{width=0.75\textwidth}
    \small
      \begin{tabular}{l|c|c|c|c}
      \hline
      \noalign{\smallskip}
      Method & Quadrant & Colour & Shape & Size \\
      \noalign{\smallskip}
      \hline
      \noalign{\smallskip}
      Baseline                      & $29.3_{0.7}$ & $28.5_{0.2}$ & $27.6_{0.2}$ & $27.3_{1.2}$ \\
      FiLM                          & $48.7_{2}$ & $89.8_{2}$ & $87.7_{1.2}$ & $84.2_{2}$ \\
      Guiding Block                 & $81.9_{3.3}$ & $89.3_{5}$ & $89.9_{2}$ & $84.3_{3}$ \\
      \hline
      Single Attention (w/o FiLM)    & $50.2_{0.7}$ & $81.5_{7.5}$ & $79.2_{6.9}$ & $80.8_{2.5}$ \\
      Multiple Attentions (w/o FiLM) & $82.4_{3.8}$ & $90.1_{2.9}$ & $88.3_{5.9}$ & $\textbf{85.4}_{2.1}$ \\
      \hline
      INSIDE (Single Attention)     & $77.9_{2.5}$ & $87.8_{0.6}$ & $89.8_{0.7}$ & $\textbf{85.4}_{0.8}$ \\
      INSIDE (Multiple Attentions)  & $\textbf{85.7}^{**}_{2.5}$ & $\textbf{90.7}^{**}_{1.9}$ & $\textbf{90.7}^{**}_{1.9}$ & $\textbf{85.4}^{**}_{1.2}$ \\
      \hline
      \end{tabular}
    \end{adjustbox}
  \end{center}
\end{table}

\subsection{ACDC Dataset}
The ACDC dataset~\cite{bernard2018deep} contains cine-MR images of 3 cardiac
structures, the myocardium and the left and right ventricular cavities, with the
task to segment these anatomies. The annotated dataset contains images at
end-systolic and -diastolic phases from 100 patients, at varying spatial
resolutions. We resample the volumes to the common resolution of 1.37 mm$^2$ per
pixel, resize each slice to $224 \times 224$ pixels, clip outlier intensities
within each volume outside the range $[5\%, 95\%]$, and finally standardise the
data to be within the [0, 1] range.

\subsubsection{Conditioning.} We evaluate two conditioning scenarios: slice
position and phase. Slice position is normalised between $[0, 1]$ (from apical
slice to basal slice), and cardiac cycle phase, i.e. end-systolic or -diastolic,
is encoded as a one-hot vector.

\subsubsection{Network.} To segment the images, we train a
U-Net~\cite{ronneberger2015u} network with 4-down and 4-up sampling blocks,
Batch Normalisation~\cite{ioffe2015batch} and ReLU activations, with a softmax
for final classification. The architecture selection was motivated by its
state-of-the-art results on the ACDC dataset~\cite{bernard2018deep}. The
conditional layers are placed along the decoding path between consecutive
convolutional blocks (each stage has two convolutional blocks). The diagrams
can be found in the supplemental material.

\begin{figure}[b!]
  \centering
  \includegraphics[width=\textwidth]{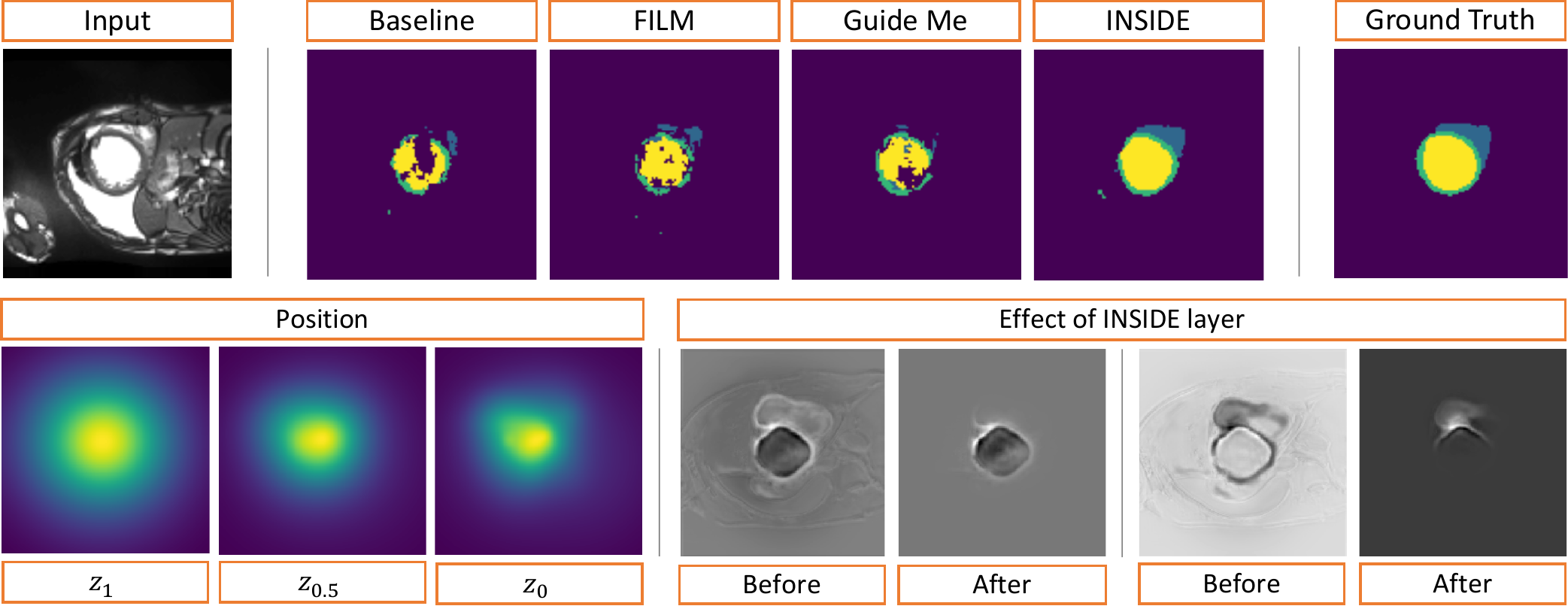}
  \caption{\textbf{Top:} Segmentation produced on
           25\% dataset, conditioning on slice position ($z$-axis). \textbf{Bottom-left:}
           Visualisation of the Gaussian attentions (averaged across channels)
           from the final INSIDE layer. Moving from the
           basal slice (denoted as $z_1$) along the $z$-axis to the apical
           slice (denoted as $z_0$), we observe the spread of the Gaussians to
           contract. \textbf{Bottom-right:} Visualisation of applying INSIDE layer.}
  \label{fig:acdc-segmentations}
\end{figure}

\begin{table}
  \begin{center}
    \caption{Quantitative results on ACDC dataset where each image is provided
             with slice position within the heart volume or phase (end-systolic or
             end-diastolic). We report average Dice score (multipled by 100) across all anatomical
             structures evaluated on whole volumes. The method with highest Dice score
             is presented in \textbf{bold}. We use the Wilcoxon test to assess statistical
             significance between INSIDE and the second best method. We denote one (*) and
             two (**) asterisks for $p \leqslant 0.1$ and  $p \leqslant 0.01$ respectively.
             We apply Bonferroni correction ($m = 2$) to account for multiple comparisons
             (when comparing to ``Baseline'').}
    \begin{adjustbox}{width=0.75\textwidth}
    \small
      \begin{tabular}{l|c|c||c|c||c|c}
      \hline\noalign{\smallskip}
      \multirow{2}{*}{Method} & \multicolumn{2}{c||}{100\% dataset} & \multicolumn{2}{c||}{25\% dataset} & \multicolumn{2}{c}{6\% dataset} \\
                              & Position & Phase & Position & Phase & Position & Phase \\
      \noalign{\smallskip}
      \hline
      \noalign{\smallskip}
      Baseline      & \multicolumn{2}{c||}{$87_{1.9}$} & \multicolumn{2}{c||}{$78.2_{2.5}$} & \multicolumn{2}{c}{$53.6_{6}$} \\ 
      \noalign{\smallskip}
      FiLM          & $84.7_{4.5}$ & $85.4_{3.9}$ & $73.4_{16.3}$ & $76_{5.4}$ & $59.9_{5.4}$ & $55.7_{7.6}$ \\
      Guiding Block & $83.7_{4}$ & $76.2_{21.4}$ & $77.8_{2}$ & $77.6_{2.4}$ & $53.7_{5.5}$ & $54.8_{5}$ \\
      INSIDE        & $\textbf{87.8}^{**}_{1.5}$ & $\textbf{87.7}^{*}_{1.6}$ & $\textbf{80.2}^{**}_{1.4}$ & $\textbf{78.9}^{*}_{1.8}$ & $\textbf{62.2}^{*}_{5.2}$ & $\textbf{61.4}^{**}_{3.8}$ \\
      \hline
      \end{tabular}
    \end{adjustbox}
  \end{center}
\label{tab:acdc}
\end{table}

\subsubsection{Results.} The empirical results on the ACDC dataset are presented
in Table 2, with varying fractions of the training set, i.e. at 100\%, 25\%, and
6\%. Overall, our method achieves consistent improvement over the baseline (when
no conditioning information is provided) with better relative performance as the
size of the training dataset decreases (+0.9\%, +2.6\%, +16\% Dice
respectively). We argue the networks rely more on non-imaging information when
the number of training examples is reduced. We further present selected
segmentation results and visualise how the attention changes depending on the
conditioning information (Fig.~\ref{fig:acdc-segmentations}). Across the
evaluation scenarios, we notice conditioning on slice position generally yields
the highest improvement, which is expected as there is a clear link between
position and heart size (expressed by the Gaussian's standard deviation). The
proposed method achieved the highest average Dice score across all tested
scenarios. The Guiding Block underperforms in most experiments, and we argue
that our choice of Gaussian attention imposes a beneficial shape prior for the
heart. FiLM performs well (at 6\% training data) when position is provided as
the conditioning information, but underperforms comparing to our method; this is
logical since non-imaging information helps to inform the network about the
expected size of segmentation masks, but does not have enough flexibility to
spatially manipulate features maps as our method.

\section{Conclusion}
Endowing convolutional architectures with the ability to peruse non-imaging
information is an important problem for our community but still remains
challenging. In this work, we have proposed a new conditional layer which extends
FiLM with Gaussian attention that learns spatial dependencies between image
inputs and non-imaging information when provided as condition. We have shown the
attention mechanism allows spatial-dependency to be modelled in conditional
layers. Our method is low in parameters, allowing efficient learning of
feature-wise attention mechanisms which can be applied to 3D
problems by adding an additional orthogonal 1D Gaussian for each channel.

\section*{Acknowledgments}
This work was supported by the Engineering and Physical Sciences Research
Council [grant number EP/R513209/1]; and Canon Medical Research Europe Ltd. S.A.
Tsaftaris acknowledges the support of the Royal Academy of Engineering and the
Research Chairs and Senior Research Fellowships scheme.

\bibliographystyle{splncs04}
\bibliography{references}

\newpage
\section*{Supplemental Material}

\begin{figure}
  \centering
  \includegraphics[width=\textwidth]{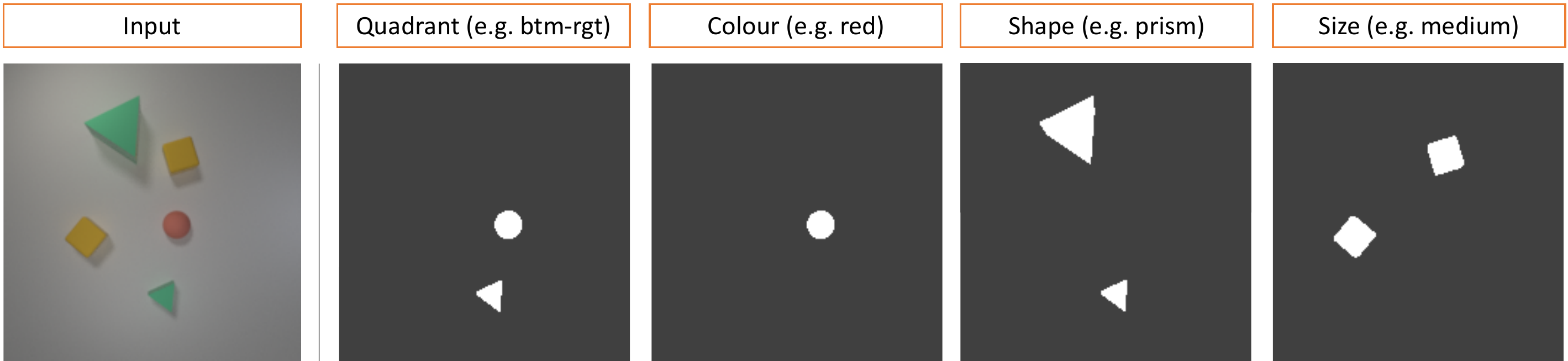}
  \caption{Different conditioning scenarios available in the CLEVR-Seg dataset.
           The desired segmentation result depends on the conditioning information,
           which relates to location, colour, shape and size. Our dataset is based
           on the CLEVR dataset~\cite{johnson2017clevr} which we have altered to
           include segmentation masks. Other modifications include changing the
           viewpoint (bird's-eye view) to avoid occlusion, replacing the cone with
           a triangular prism (the cone's circular base might be confused with a
           sphere), removing the reflective texture and finally resizing the images
           to $200 \times 200$ pixels.}
  \label{fig:clevr}
\end{figure}

\begin{figure}
  \centering
  \subfloat{%
  \includegraphics[width=0.75\textwidth]{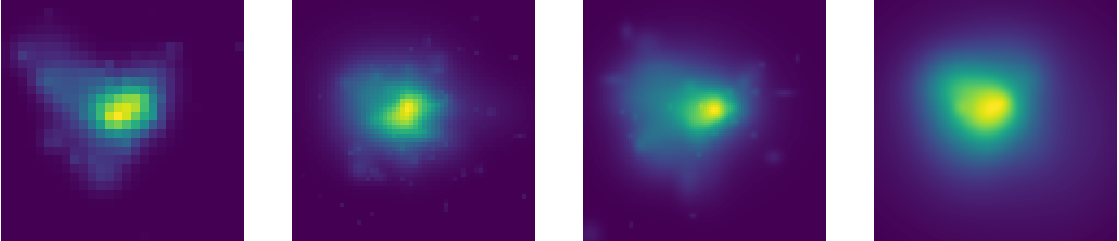}
  }\hfil
  \subfloat{%
  \animategraphics[width=0.161\textwidth,loop,controls=play]{12}{figures/gif/att_}{0}{9}
  }
\caption{\textbf{Left:} Visualisation of Gaussian attentions (averaged across
         all channels) at different up-sampling stages (from the smallest
         resolution to the largest) conditioned on apical slice position.
         \textbf{Right:} Animated visualisation (requires Adobe Acrobat to play)
         of the final INSIDE layer interpolated across all slices (from basal to
         apical).}
\end{figure}

\begin{figure}
  \centering
  \includegraphics[width=\textwidth]{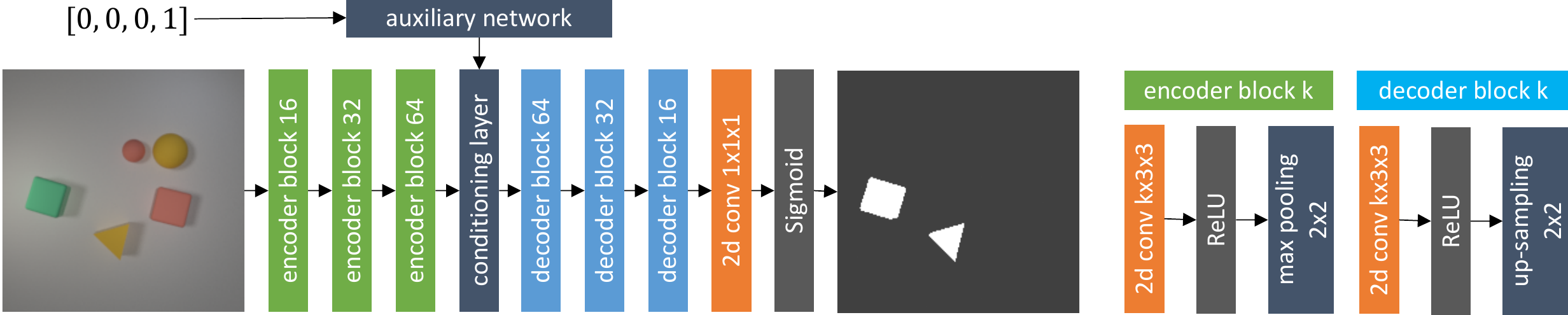}
  \caption{Experiment setup for the CLEVR-Seg dataset. All conditioning
           mechanisms have been evaluated using the same backbone network.
           We use an auxiliary network to predict parameters for the
           conditional layers arranged as
           ($64 - 64 - n$) with $n = 128$ for FiLM, $n = 228$ for Guiding Block,
           and $n = 384$ for INSIDE. We use $\tanh$ activation function in every layer except the final layer.}
  \label{fig:encoder-decoder}
\end{figure}

\begin{figure}
  \centering
  \includegraphics[width=\textwidth]{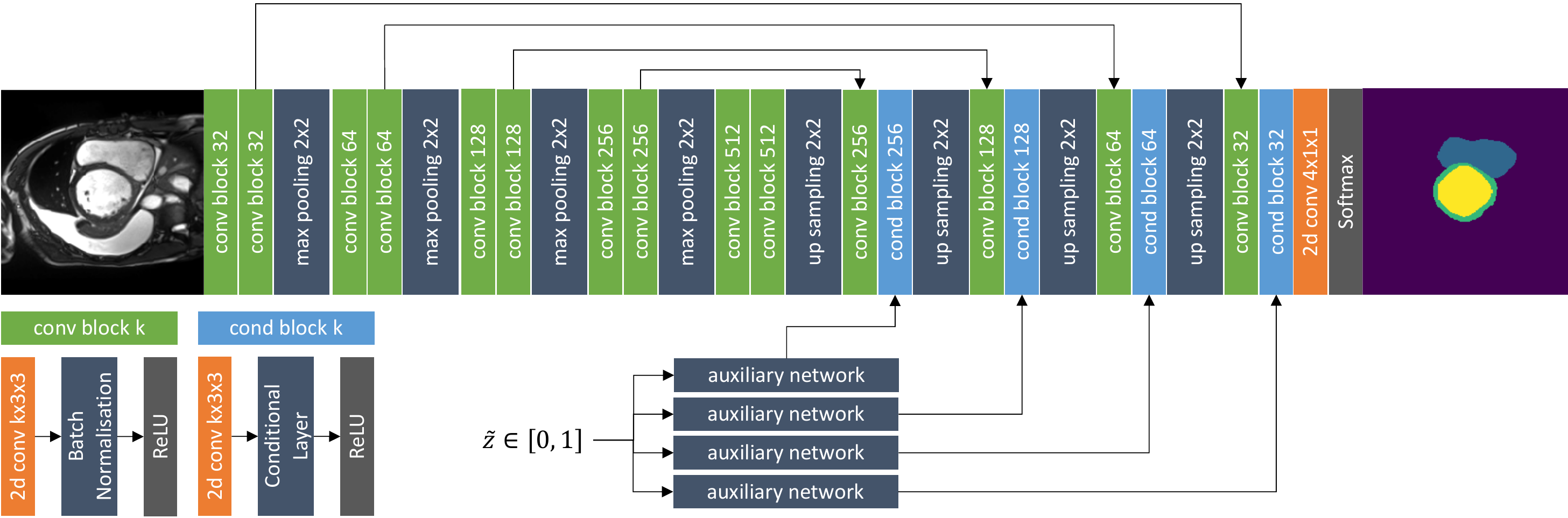}
  \caption{Experiment setup for the ACDC dataset. Our network follows the U-Net architecture~\cite{ronneberger2015u} with skip-connections
           (arrows above). We use separate auxiliary networks for each layer
           following a ($\frac{c}{2} - \frac{c}{2} - n$) architecture where
           $c$ is the number of channels. We place FiLM, Guiding Block,
           and INSIDE at the point where we indicate \emph{conditional layer}. The baseline
           uses Batch Normalisation~\cite{ioffe2015batch} instead of a
           \emph{conditional layer}.}
  \label{fig:unet}
\end{figure}

\end{document}